\documentclass[letterpaper, 10 pt, conference, hyphens]{ieeeconf}  % Comment this line out
\IEEEoverridecommandlockouts                              % This command is only
\usepackage[utf8]{inputenc}
\usepackage[T1]{fontenc}
\usepackage{amsmath}
\usepackage{amssymb}
\usepackage{bm}
\usepackage{epsfig}
\usepackage{graphics}
\usepackage{mathptmx}
\usepackage{hyperref}
\usepackage{mathtools}
\usepackage{xcolor}

\begin{document}

\title{\LARGE \bf A Low-Cost Portable Apparatus to Analyze Oral Fluid Droplets and Quantify the Efficacy of Masks\\
\vspace{5mm}

\normalsize
Ava Tan Bhowmik\\
The Harker School, San Jose, CA \\
ava.t.bhowmik@gmail.com
\vspace{-1.4cm}
}

\maketitle

\begin{abstract}

Every year, about 4 million people die from upper respiratory infections. Mask-wearing is crucial in preventing the spread of pathogen-containing droplets, which is the primary cause of these illnesses. However, most techniques for mask efficacy evaluation are expensive to set up and complex to operate. In this work, a novel, low-cost, and quantitative metrology to visualize, track, and analyze orally-generated fluid droplets is developed. The project has four stages: setup optimization, data collection, data analysis, and application development. The metrology was initially developed in a dark closet as a proof of concept using common household materials and was subsequently implemented into a portable apparatus. Tonic water and UV darklight tube lights are selected to visualize fluorescent droplet and aerosol propagation with automated analysis developed using open-source software. The dependencies of oral fluid droplet generation and propagation on various factors are studied in detail and established using this metrology. Additionally, the smallest detectable droplet size was mathematically correlated to height and airborne time. The efficacy of different types of masks is evaluated and associated with fabric microstructures. It is found that masks with smaller-sized pores and thicker material are more effective. This technique can easily be constructed at home using materials that total to a cost of below \$60, thereby enabling a low-cost and accurate metrology.
\end{abstract}

%%%%%%%%%%%%%%%%%%%%%%%%%%%%%%%%%%%%%%%%%%%%%%%%%%%%%%%%%%%%%%%%%%%%%%%%%%%%%%%%
\section{\textbf{INTRODUCTION}}

The COVID-19 pandemic is the most devastating global health crisis since the 1918 influenza pandemic [1-4]. As of December 2022, the SARS-CoV-2 virus has infected over 656 million people and caused 6.67 million fatalities worldwide [5]. Besides COVID-19, the influenza virus infects approximately one billion people worldwide and inflicts close to 650 thousand deaths annually [6]. Globally, respiratory syncytial virus (RSV) infects 64 million people and causes 160,000 deaths each year [7]. Contagious respiratory diseases spread through the corresponding pathogens in saliva or mucus droplets generated by infected individuals when they breathe, talk, cough, or sneeze. Inhalation of such particles often causes infection [8]. However, the propagation of these infectious droplets can be obstructed by wearing an effective mask [9]. Proper mask usage could potentially prevent 4 million acute upper respiratory infection-induced deaths each year [10].

Standard masks such as N95 respirators are proven to be the most efficient at filtering orally-generated fluid droplets, blocking 95\% of airborne particles [11]. Top-grade personal protective equipment (PPE) should be prioritized for medical first responders and other healthcare workers, especially amidst PPE shortages [12]. Additionally, surgical and N95 masks are single-use and are made out of non-recyclable plastics that are harmful to the environment when disposed of [13-15]. Although KN-95 masks have recently become prevalent, many people are not wearing them correctly or frequently [16]. As a result, the general public is encouraged by the Centers for Disease Control and Prevention (CDC) to wear alternate face coverings such as homemade or commercially available fabric masks [17]. However, many such masks are ineffective and there is no easy and low-cost method for determining the efficacy of specific, individual masks accurately. Quantitative experimental setups that have been reported require a laboratory, complex equipment, and experience to operate. They are generally not accessible and repeatable in home environments. It is both costly and impractical to test every single kind of mask in a lab setting. A lack of accessible methods to verify mask efficacy may result in  usage of masks that do not effectively block droplets unknowingly. 

In this work, a novel, low-cost, portable, and accurate apparatus to visualize orally generated fluid droplets and quantify mask efficacy has been developed. This method is the world’s first fluorescence-based technique for oral droplet visualization utilizing common household materials. It is based on the fluorescent properties of tonic water, a common beverage, which makes the  oral droplets generated by expiratory events visible under UV darklight. The fluorescent emission is captured in slow-motion using a smartphone camera and processed and analyzed using open-source software. Detected droplet size was estimated based on the test conditions and the flight time of the droplets with a corresponding mathematical model. The results of these analyses are applied to determine the sensitivity of the metrology and the efficacy of masks.

\section{\textbf{PRIOR RESEARCH}}

Past reports on oral fluid droplet visualization require advanced equipment and are complex to operate. Representative approaches regarding the detection and/or visualization of respiratory droplets are summarized below.

Bourouiba et al. utilizes light scattering with high-speed videography to visualize the cloud of saliva and mucous droplets expelled by a sneeze [18]. Saliva droplets propel from an individual’s mouth or nose scatter photons, making them visible against a black background. This method is accurate and can capture fast-moving droplets, but other foreign particles present in the test chamber would also scatter light, resulting in unwanted signal. To address this problem, a high-efficiency particulate air (HEPA) filter is used. However, the filtration system still allows some dust particles to enter the test chamber [19]. In addition, a high-power light source is needed to generate sufficient signal from the droplets and the wavelength of light scattered by the droplets is the same as the wavelength of the source light, which reduces the signal-to-noise ratio (SNR) for detection. A better way to avoid this specific issue is to utilize fluorescence, where the emission wavelength is lower than the excitation wavelength [20].

Another type of experiment utilizes a laser light sheet rather than a large-area light illumination to avoid the source light being captured by the camera and increasing the background noise [21]. However, this technique can only visualize droplets passing through the thin light sheet at a single given moment. Furthermore, if not handled properly, the powerful laser can cause eye damage and visual impairment if not handled properly [22]. In addition, the expensive equipment such as HEPA filters, high-intensity lasers, and high-speed cameras render these experiments inaccessible in home environments.

To lower the overall cost of mask efficacy evaluation, several low-cost techniques could potentially be used for testing in a home setting. Unfortunately, these simple methods for mask efficacy determination, such as the “candle flame test” [23] or hydrophobic coating test are not quantitative and lack accuracy. In the example of the candle test, in which a subject attempts to blow out a candle while wearing a mask, outside variables, such as the type of candle and personal lung strength can affect the outcome. In addition, blowing may not be a reliable proxy for small aerosols exiting with normal speaking or coughing. In the hydrophobic coating test, the presence of a hydrophobic covering may not necessarily correlate to mask efficacy, as factors like pore size and fabric layering effect mask efficacy too. One other test method that is potentially feasible at home uses colored dyes to quantify the number of droplets generated by an expiratory event [24]. This system has several limitations: The microdroplets that are less than 10 microns in diameter cannot be tracked and recorded while airborne due to the low SNR and high viscosity of droplets, thus eliminating the possibility of studying the trajectory and propagation of the particles. The dye droplets leave a distorted mark when landing on the ground, causing some stains to merge and ruining the analysis. Additionally, the dyes could potentially be toxic to ingest, unpleasant to taste, and difficult to clean up.

\section{\textbf{METROLOGY DEVELOPMENT}}

The goal of the metrology development phase was to leverage common household items for quantitative and accurate droplet detection, use smartphone-based high-speed videography, and apply open-source software for image processing and analysis. This development project consists of four major modules: setup optimization, data collection, data processing and analysis, and application development, as illustrated in \hyperref[fig1]{Figure 1}. For setup optimization, various fluorescent liquids, UV light sources, and data collection settings were characterized and the optimal combination of conditions determined. During data collection, recordings of the fluorescent microdroplets are captured with a smartphone. The data is processed and analyzed using an automated macro. This metrology can be applied to a broad range of applications, including the study of droplet generation and propagation, and mask efficacy testing, and has been built into a portable prototype for education purposes to demonstrate the spread of respiratory diseases. Details for each module are described in the following sections.

\begin{figure*}[hbt!]
      \centering
      \label{figure1}
      \includegraphics[scale=0.34]{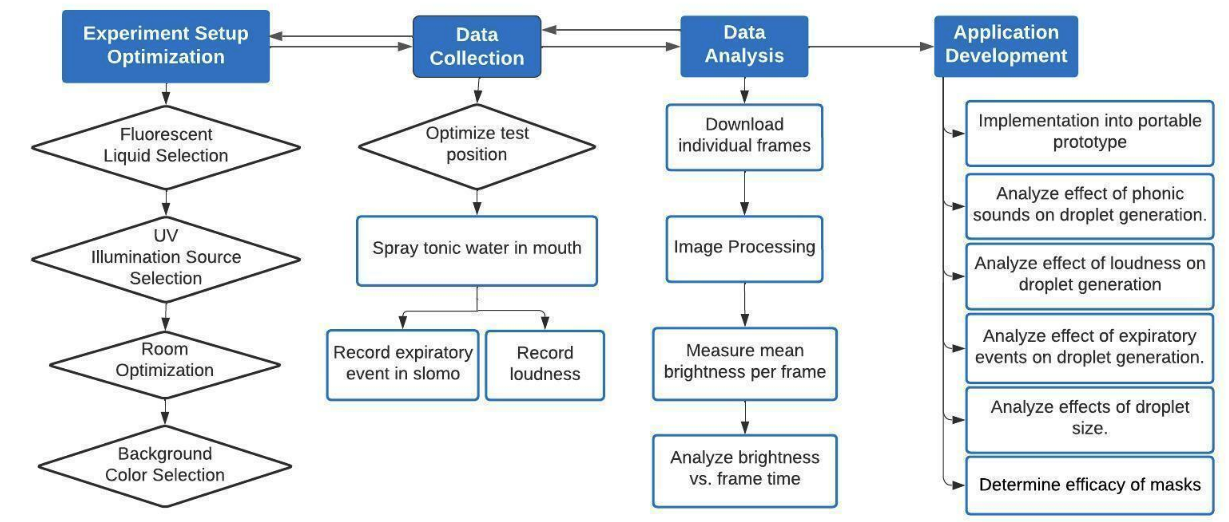}
      \caption{The flowchart of the metrology development process. The four modules are setup optimization, data collection, data analysis, and application development. }
\end{figure*}

\subsection{Setup and Material Optimization}

During the setup and material optimization phase of the project, independent variables such as fluorescent liquid choices, UV light source selection and configuration, and test setup conditions were evaluated and determined. For each variable, background research, comparisons, multiple testings, and analysis were performed.

\subsubsection{Fluorescent Liquid Selection}

The first step of the research is to select an ingestible, fluorescent liquid that has a similar viscosity to saliva and fluoresces brightly to be detected and captured using an iPhone camera. This liquid is used as a proxy for oral fluid droplets. After studying a wide range of common household products, tonic water was the only liquid that fulfilled all the criteria. Tonic water contains a fluorophore called quinine [25]. The excitation wavelength range of quinine is ~270 to 400 nm and the emission spectrum is ~380 to 530 nm, giving tonic water its signature blue glow under UV dark light [26].

To optimize fluorescent intensity, various quinine-containing liquids and concentrations were evaluated. Figure 2 shows comparison images of the test liquids with tap water as a control. A notable candidate was East India Tonic Syrup, advertised to contain 5 times the quinine concentration of tonic water. However, as is apparent in Figure 2A, the Schweppes tonic water was significantly brighter than the tonic syrup. This is because the tonic syrup contains cinchona bark, from which quinine is extracted, rather than pure quinine. The impact of tonic water concentrations were tested, including 33\%, 50\%, 100\%, and 200\% by dilution or evaporation. The results are shown in Figure 2B. Although evaporating tonic water to obtain a higher concentration resulted in a higher brightness level than regular tonic water, regular Schweppes tonic water was found to be the most suitable for the experiment since it introduces the least variability without risking possible quinine decomposition from high temperature [27].

\begin{figure*}[thpb]
      \centering
      \label{figure2}
      \vspace{1 cm}
      \includegraphics[scale=0.39]{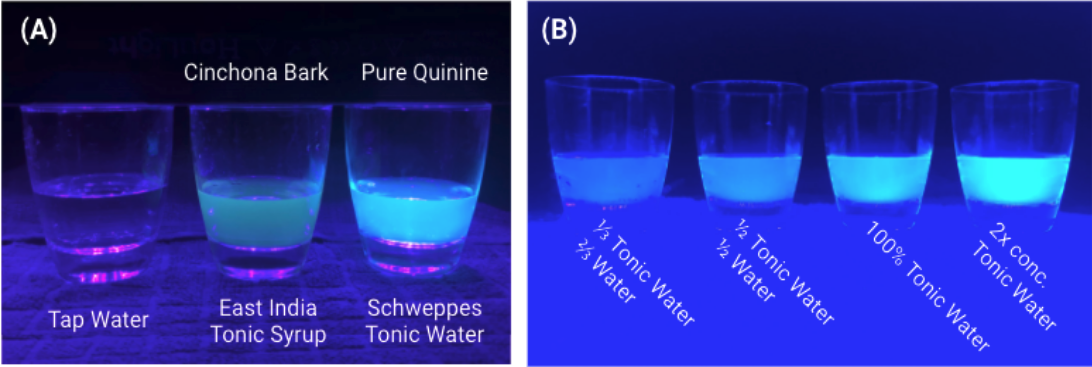}
      \caption{Comparison of the fluorescence intensity of (A) various quinine-containing liquids with tap water as a control, (B) different concentrations of tonic water under UV illumination.}
\end{figure*}

\subsubsection{UV Light Source Selection}
Both a UV blacklight flashlight (385-395 nm wavelength) and UV darklight party tube lights (397-402 nm wavelength) with different power settings and configurations were tested. The tube light was determined to be more suitable than the flashlight since it provides uniform intensity over the illumination field, whereas the flashlight has a radial intensity decay towards the edge of the field. When a flashlight is used, only droplets passing through the center of the light beam are visible. Additionally, considering safety, the party tube light had a wavelength that was safe for exposure to human eyes and skin [28].

\subsubsection{UV Light Source Selection}
Figure 3 shows a matrix of different setup conditions experimented with to determine the optimal configuration for final testing. This includes the color of the background and the state of the room light. A spray bottle is used during these calibration tests to ensure low variability in the process of droplet generation. It is determined that data collected with a black background in a dark room was the most optimal. This is because the black background absorbs the incoming UV light rather than reflecting it. Along with the removal of the background room light, these settings had the lowest noise and resulted in an increased SNR. 

\begin{figure*}[thpb]
      \centering
      \label{figure3}
      \includegraphics[scale=0.4]{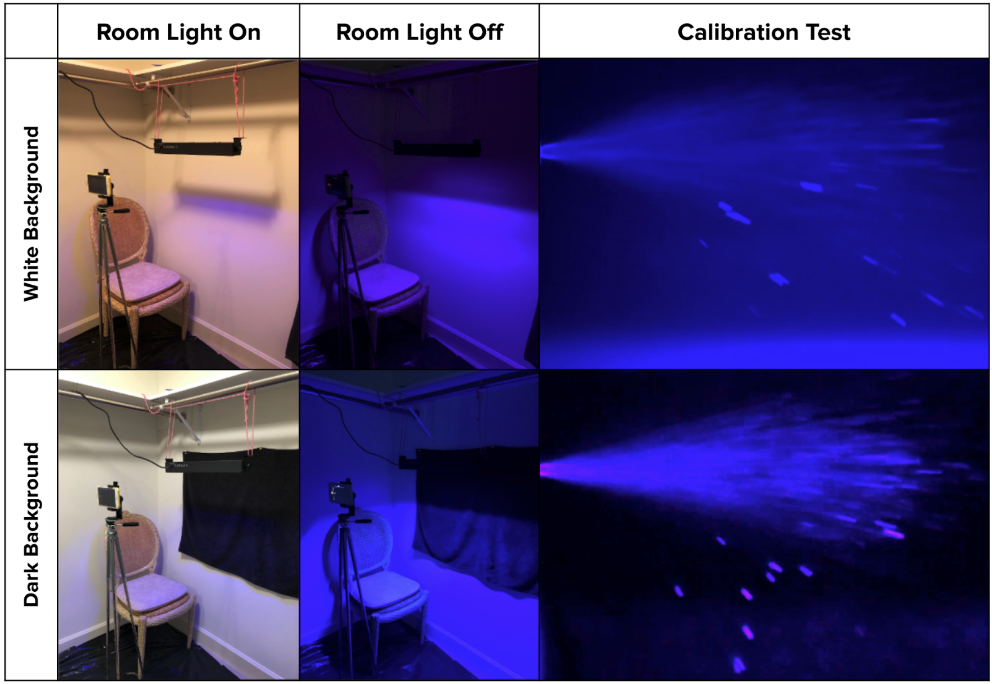}
      \caption{Comparison of different test setup conditions and their impact on the results. The independent variables changed are the color of the background and the state of the room light.}
\end{figure*}

\subsubsection{Final Metrology Setup}
Figure 4 shows the final metrology development setup constructed in a closet as the proof of concept. Besides the conditions outlined in Section 3.1.3, several other conditions are optimized. It is found that since the light field from the tube light disperses away in a trapezoid shape, the tube lights should hang at least 14 inches away from the back wall to minimize UV light reflection appearing in the video. The ideal horizontal distance between the iPhone camera and the UV tube lights is around 16 inches. At this distance, the camera can capture the entire length of the UV tube lights in the field of view to track the trajectory of the droplets without having to digitally zoom in, which can potentially reduce image quality. The optimized vertical distance between the user’s mouth and the UV tube light is 6 inches. This ensures that the droplet cloud is as close to the light as possible for the highest illumination without being obstructed.

\begin{figure*}[thpb]
      \centering
      \label{figure4}
      \includegraphics[scale=0.46]{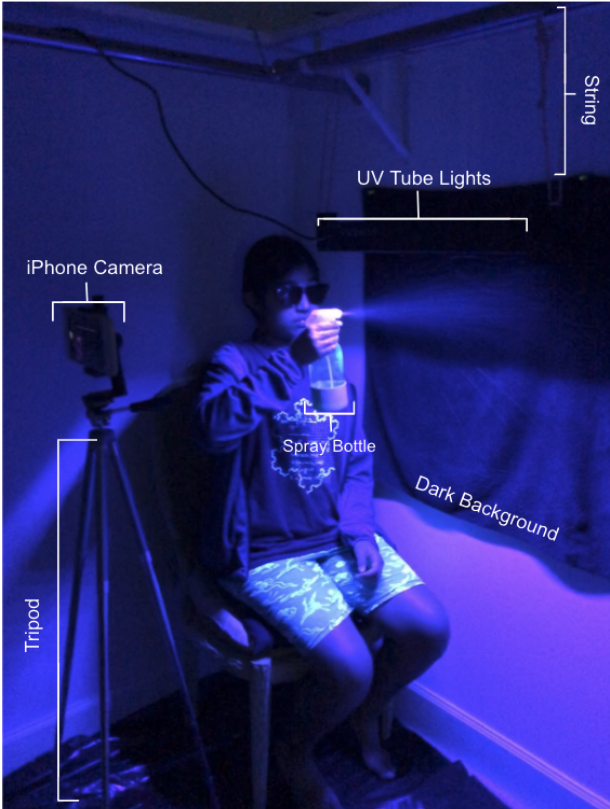}
      \caption{The final experimental setup and materials used for the metrology development.}
\end{figure*}

As seen in Figure 4, the final setup consists  of UV tube lights suspended from the closet hanger rod 14 inches away from the back wall, 16 inches away from the camera, and 6 inches below the source of the droplets, an iPhone camera placed on a tripod, a spray bottle filled with quinine-containing tonic water for calibration, and a black poster board or towel to serve as the dark background. 

\subsection{Data Collection}
During data collection, the test subject first wets their mouth with tonic water with a spray bottle to reduce variability of tonic water volume. Then, the person performs an expiratory event (speaking, sneezing, or coughing). For each trial, the loudness level of speech is measured by Starkey’s SoundCheck App and recorded. When equivalent conditions are compared, the loudness level is calibrated to remain constant. The iPhone camera is set to slow-motion mode (“slow-mo”) and used for recording at 240 frames per second (fps). Using this process, the droplets generated are visualized, recorded, and analyzed as described in Section 3.3.

\subsection{Data Analysis}
The video of the aerosol cloud generated by an expiratory event is downloaded into separate frames using the open-source VLC software package [29]. This is performed using VLC’s scene video filter, selectable under the “Preferences” menu. The recording ratio should be set to 1 to ensure that no compression is applied and every single frame in the video is saved. Next, the downloaded images from each frame are imported into ImageJ from the National Institutes of Health (NIH) [30] as an image stack for processing and analysis. In ImageJ, the frames can be enhanced by histogram adjustment. The enhanced images can be seen in the image montage shown in Figure 5. The montage shows the progression of the microdroplet cloud generated by the word “Fruits” being spoken at 92 decibels (dB) ± 3 dB. This technique can capture the microdroplets lingering in the air long after their dispersal. This shows that 6 feet of social distancing alone is not enough to combat COVID-19, as the contagious particles can stay in the air long after a sick individual has left the area. 
For quantitative analysis, the mean brightness values of each frame are collected using the “Measure…” function on the ImageJ software menu. The mean brightness values can then be graphed in an intensity vs. time plot to quantify and visualize the generation and dissipation of droplets and aerosols, as narrated in the Results and Discussion section below.
This entire analysis process is automated by coding a macro in ImageJ, which reduces the analysis time from 30 minutes to ~30 seconds.

\begin{figure*}[thpb]
      \centering
      \label{figure5}
      \includegraphics[scale=0.35]{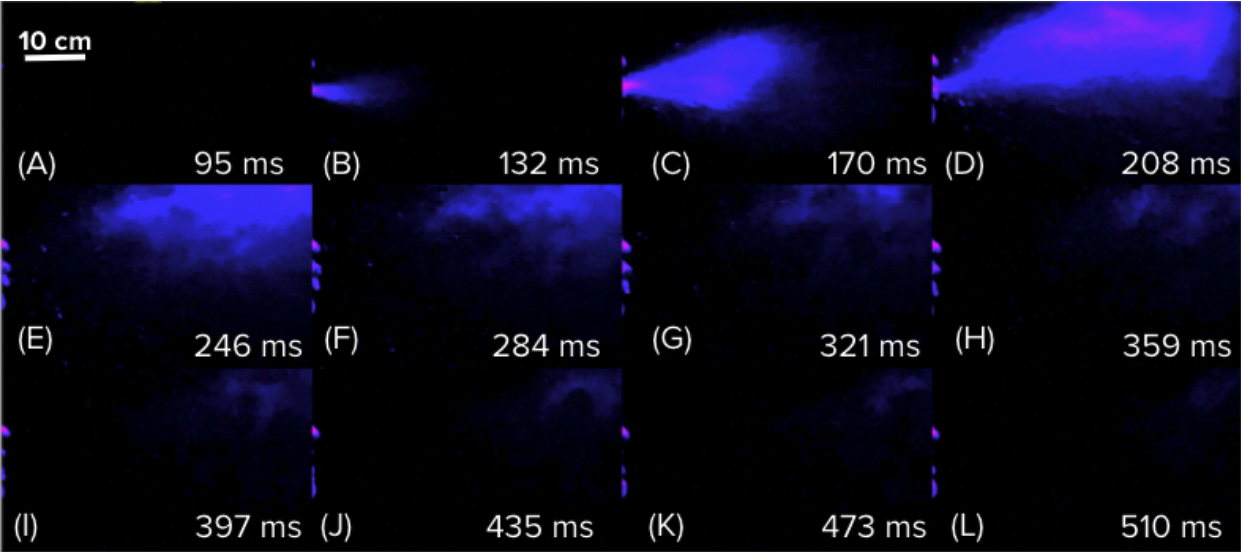}
      \caption{Propagation of microdroplets generated by “Fruits” spoken at 92 dB. The video is captured in slo-mo mode at 240 fps. The frames extracted from the video are shown in (A)-(L) with 38-millisecond intervals.}
\end{figure*}

\subsection{Applications}
Once the initial proof of concept metrology was developed in the closet, the setup was implemented into a consolidated, portable prototype in a cardboard box. This prototype can be used for mask or instrument cover evaluation purposes. Furthermore, the simple visualization of the oral fluid droplets can be used for educational purposes around the world on how contagious respiratory diseases spread through aerosols. 

\subsubsection{Portable Prototype Development and Implementation}
A portable prototype, shown in Figure 6, is constructed using a cardboard box, 2 UV tube lights, an iPhone, and black paper. The cost of the setup is less than \$60 if a phone is already in possession; the cost breakdown is shown in Table 1. 

\begin{table*}[thpb]
      \centering
      \label{table1}
      \includegraphics[scale=0.32]{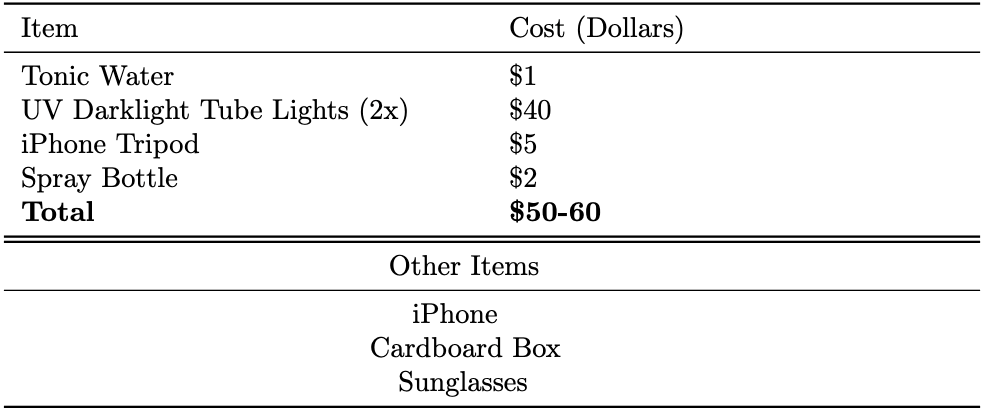}
      \caption{All materials used for setup construction and cost breakdown.}
\end{table*}

The size of the prototype displayed in Figure 6 is 38.5 inches x 35.5 inches x 28.5 inches, which can be reduced to as small as 20 inches x 30 inches x 19.5 inches. The UV tube light is placed 16 inches from the camera mounted on the front wall to ensure that the entire illumination field was captured in the field of view and 14 inches from the back to reduce the reflection of UV light. 

\begin{figure*}[thpb]
      \centering
      \vspace{0.5 cm}
      \label{figure6}
      \includegraphics[scale=0.38]{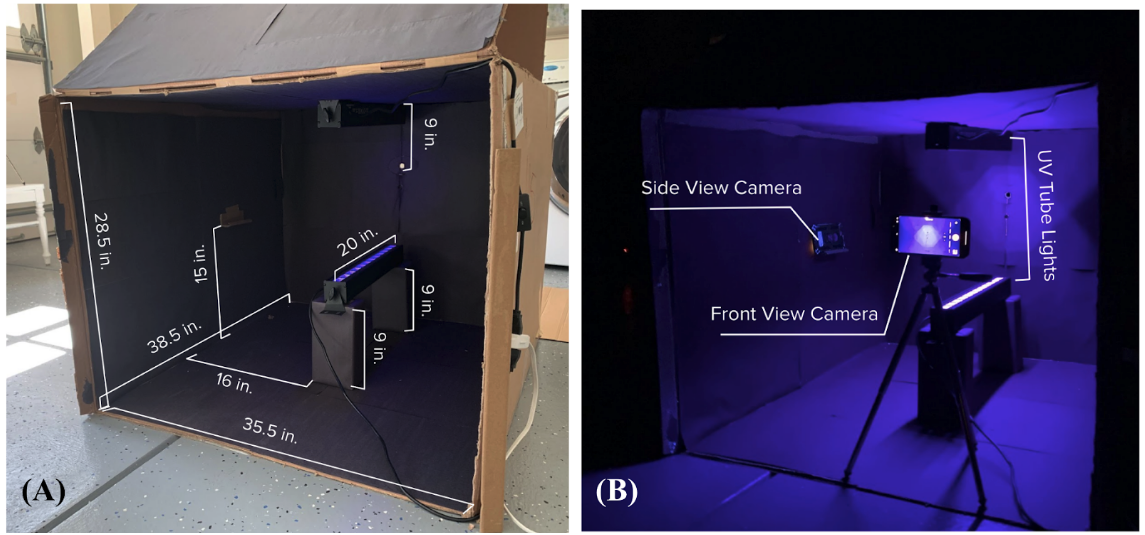}
      \caption{A miniaturized and optimized portable prototype based on the original metrology built in a cardboard box. (A) shows the setup with labeled dimensions and (B) during operation.}
\end{figure*}

\begin{figure*}[thpb]
      \centering
      \label{figure7}
      \includegraphics[scale=0.5]{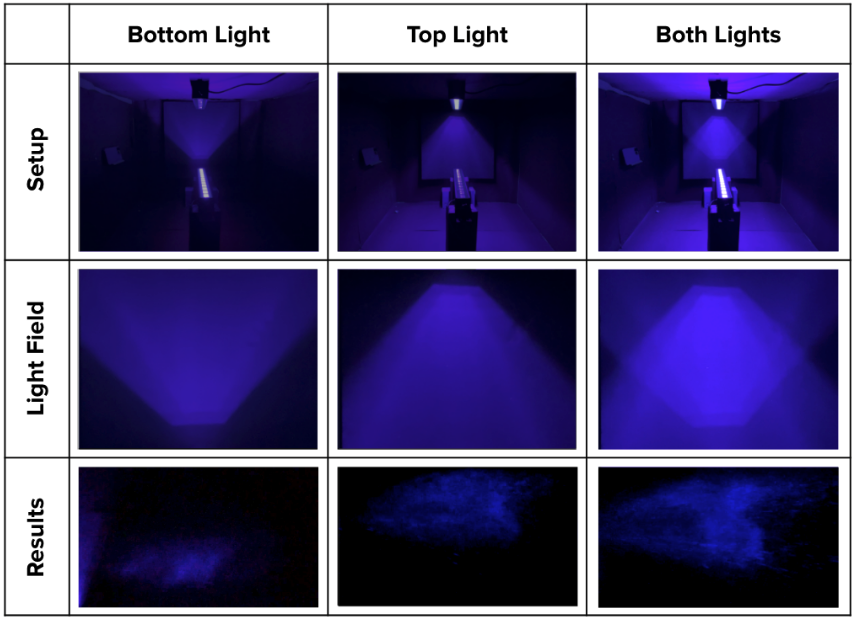}
      \caption{Comparison of the setup, light field, and results with one light versus two lights.}
\end{figure*}

\begin{figure*}[thpb]
      \centering
      \label{figure8}
      \includegraphics[scale=0.46]{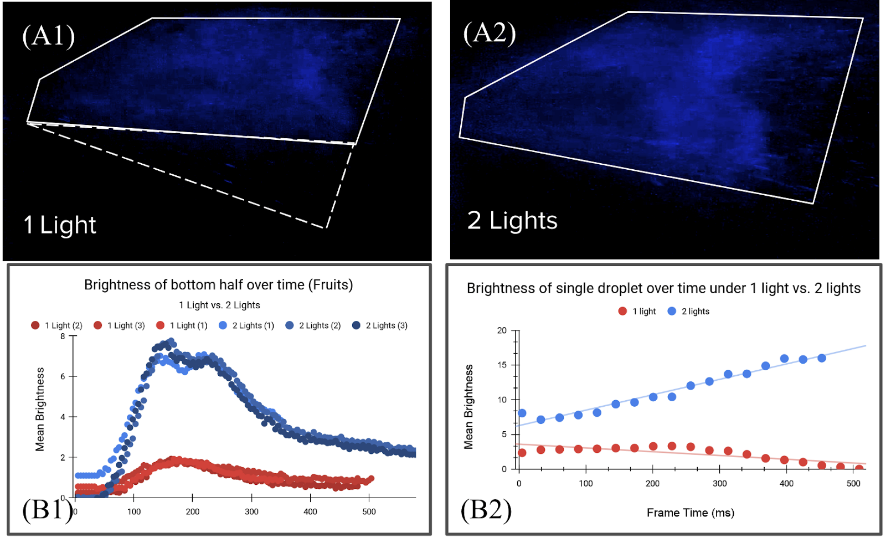}
      \caption{(A1) and (A2) show droplets captured under one and two lights. (B1) shows the brightness of the frame’s bottom half over time under the different light configurations. (B2) shows the brightness of a single droplet under the different light configurations.}
\end{figure*}

In the original single-light setup, the illumination field intensity decreases as a function of the distance from the light source, as the droplets fall further away from the light source, the bottom half of the image frame has lower illumination intensity, resulting in a decreasing signal. To further optimize the setup, a second tube light is added 20 inches away from the top light at the bottom, facing up to produce a combined, more uniform illumination field, as shown in Figure 7 and 8A. The analysis illustrates that the bottom half illumination showed a 4x increase in brightness , indicating not only an overall intensity increase, but also better signal uniformity, which can be seen in Figure 8B1. Additionally, with two lights, the droplets did not experience the same drop in intensity, as shown in Figure 8B2.

Adding a second light improves the illumination intensity by 1.5-2x, as shown in Figure 9. In this study, three words were spoken at 92 dB. There were three trials for each word spoken under 1 vs. 2 UV lights. When the words “Fruits”, “Phosphorus”, and “Philosophy” were spoken under two lights versus the one light test conditions, the intensity of the signal improved 2.1x, 1.5x, and 1.5x, respectively. It was also found that data collected using the new setup has 1-3\% lower variability across 3 trials under otherwise identical conditions.

\begin{figure*}[thpb]
      \centering
      \label{figure9}
      \vspace{1 cm}
      \includegraphics[scale=0.36]{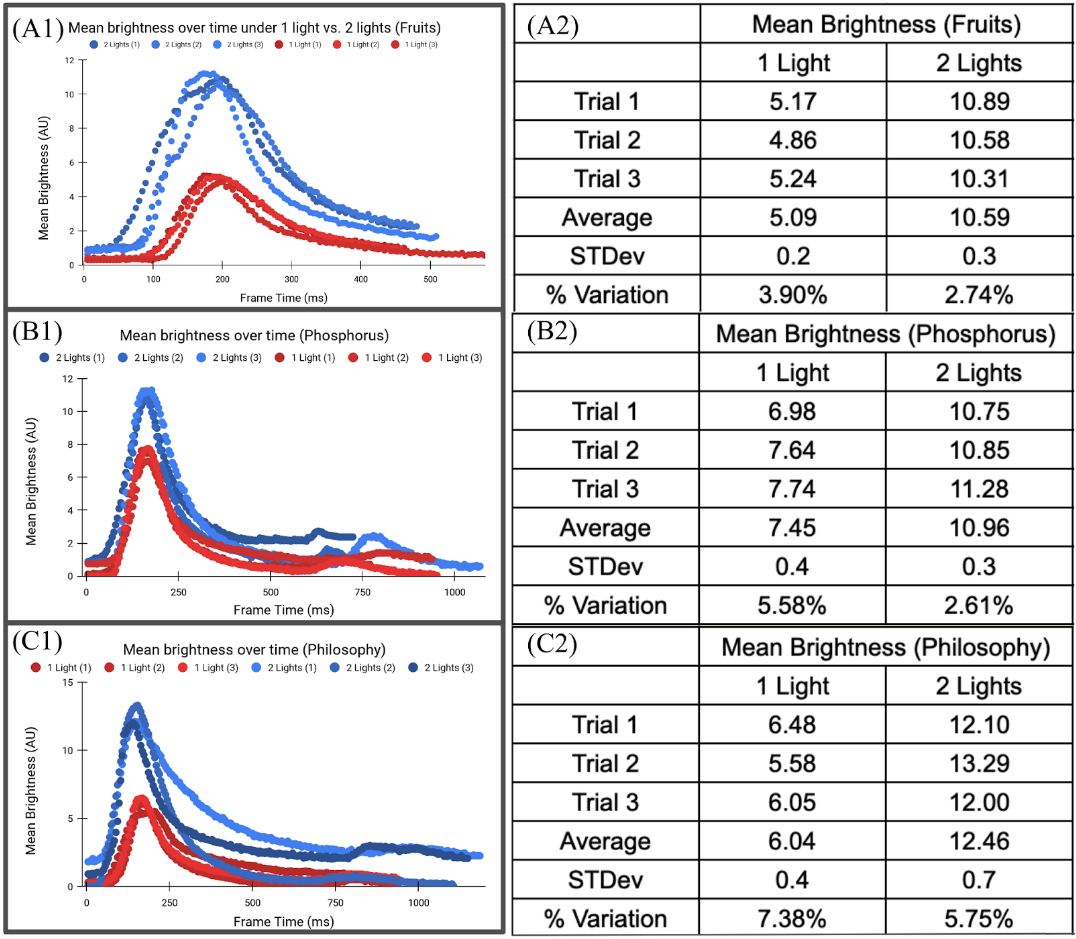}
      \caption{(A1), (B1), and (C1)’s mean brightness values over time following the utterances of the words Fruits, Phosphorus, and Philosophy. Each test condition was repeated with three trials. The red curves represent data collected with only 1 tube light; the blue curves represent those collected with 2 tube lights. (A2), (B2), and (C2) present the peak brightness values, averages, and standard deviation for each trial.}
\end{figure*}

\subsubsection{Mask Efficacy Comparison}
This portable apparatus can be applied to study the generation, trajectory, and dissipation of orally-generated microdroplets for quantitative and in-depth studying of the mechanics of droplet propagation. People use cloth masks for many different reasons: to mitigate the environmental impact of disposable masks, for comfort, or limited access to medical masks. Thus, having access to a low-cost and convenient setup as described in this work is important for individuals to evaluate whether the masks that they are using are effective. Better understanding of aerosol physics may permit specialized mask designs specifically suited towards blocking the user’s droplets from escaping when speaking, coughing, and sneezing. Furthermore, such a setup would expedite the process of developing and testing revolutionary material, such as copper foam [31], that would mitigate the environmental impact of widespread face mask use. Additionally, this setup would be useful in areas with limited resources to not only demonstrate the mechanism of virus spread, but also for the aforementioned mask evaluation purposes. 

With the prototype described above, a number of trials were repeated with the subject wearing a different mask each time. The masks tested are made of thin cotton, thick cotton, linen, thin polyester, thick polyester, surgical, and N95. The peak brightness values from the recordings of each trial for a given mask were recorded. The efficacy of each tested mask is determined and correlated to its nanostructure using averaged peak intensity values and the scanning electron microscopy (SEM) images of different fabrics, which can be seen in Figure 10. The bigger and greater density of pores there are, the more droplets escape. Thinner material is correlated with more droplets being generated. Thick and thin cotton, which are both porous and the least effective. Not only did they allow small droplets to pass through, but they also allowed larger droplets to be broken up into aerosols when they encountered the holes. Linen blocked many aerosols in addition to small and big droplets but still let a considerable amount through. Polyester barely let any droplet through and the surgical and N95 masks were very effective. 

\begin{figure*}[thpb]
      \centering
      \vspace{1 cm}
      \label{figure10}
      \includegraphics[scale=0.35]{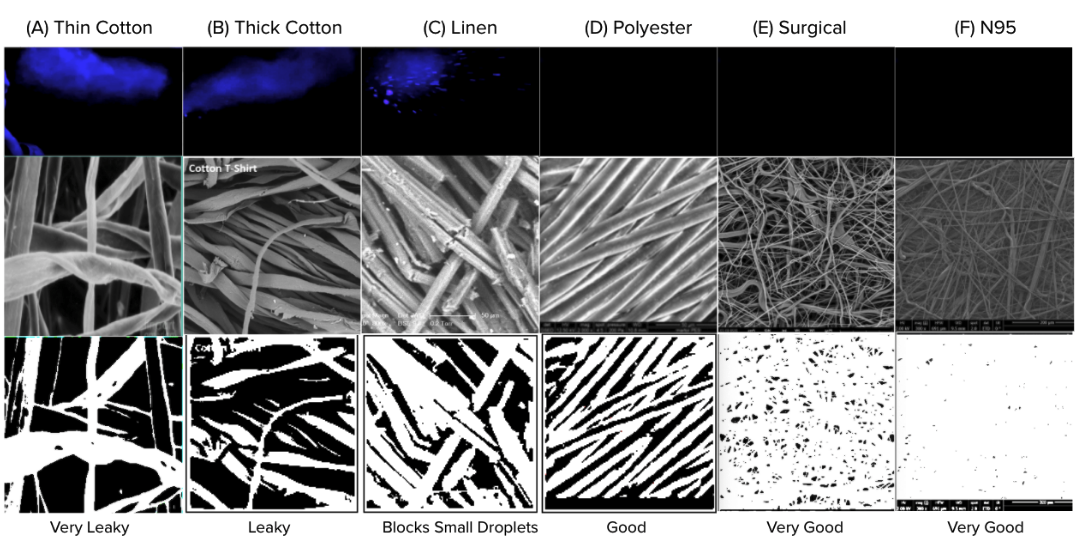}
      \caption{(A)-(F) represent different materials used in the study. Micrographs of different mask fabrics captured using scanning electron microscopy (SEM). This corresponding  brightest frame from the videos of each mask are correlated with SEM images of the fabric and the thresholded images.}
\end{figure*}

As compared in Figure 11, of all the face coverings tested (both homemade and standard), the N95 mask (Figure 10F) predictably performs the best. A thick polyester mask (Figure 10D) *is the most effective cloth mask at blocking the droplets. It can be observed that the droplets escaping from the cotton masks (Figure 10A, 10B) are significantly smaller than those with the linen face covering (Figure 10C), almost a fine mist compared to the distinctive, individual droplets. This may be due to the initial droplets passing through the holes in the cotton fabric becoming broken up into smaller droplets. However, in linen, due to the larger pore size, it is possible some droplets passing through the fabric will merge on the other side. As a result, while cotton (Figure 10A, 10B) masks can help hold back oral droplets, it is possible that wearing them may be even more dangerous than not wearing a mask at all due to droplets that would otherwise fall to the ground quickly being converted into aerosols that linger in the air.

\begin{figure*}[thpb]
      \centering
      \label{figure11}
      \includegraphics[scale=0.4]{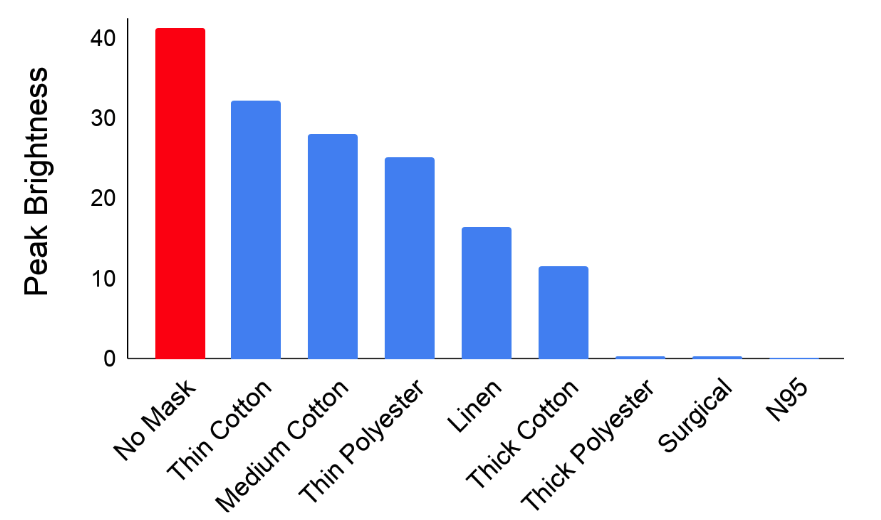}
      \caption{Comparison of the efficacy of the masks as measured by the peak brightness analyzed based on the recordings. The masks analyzed include thin cotton, medium cotton, thick cotton, thin polyester, thick polyester, linen, surgical and N95.}
\end{figure*}

\section{\textbf{DISCUSSIONS}}

Using the experimental setup, data collection, and automated analysis procedures described in Section 3, the effects of variables such as the phonic sound, loudness levels, and type of expiratory event on the amount and propagation characteristics of droplets generated were studied. In this section, the results of these analyses are discussed.

\subsection{Dependency of Droplet Propagation on Droplet Size}

From the data collected with this prototype, the size of the droplets can be modeled and correlated. The results reaffirms the already well established fact [32, 33] that aerosols, which are droplets smaller than 10 microns in diameter, float in the air for prolonged periods of time and eventually dehydrate, making them particularly infectious. The dehydrated nuclei of the aerosols can stay in the air for minutes or even hours. As a result, they are much more dangerous than larger droplets that immediately fall to the ground following expulsion. 
The mean time for a particle to reach the ground can be calculated using the equation:
\begin{figure}[hbt!]
\centering
\includegraphics [scale=0.34]{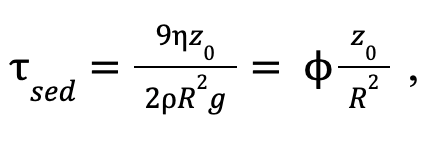}
\end{figure}

where $\tau_{sed}$ is the time that it takes for a droplet of radius R to reach the ground from a height, $z_0$ (in micrometers) [34]. The prefactor, $\phi = 9\eta/(2\rho g) = 0.85 * 10^{-2}\mu m*s$, is calculated based on the viscosity of air at 25 $^{\circ} C$, $\eta = 1.86 * 10^{-8} g/(\mu m*s)$, water density $\rho = 10^{-12} g/\mu m^3$, and the gravitational constant $g = 9.8 * 10^6 \mu m/s^2$. Using this equation, the time it takes for a droplet to fall to the ground in the absence of evaporation can be calculated based on the size of the droplet. For example, droplets placed initially at $z_0 = 1.5 m$ (the average height above ground for the mouth of a standing human adult) with radii of 1, 10, or 100 $\mu m$ will require $1.3 * 10^4 s (3.5 hrs)$, $130 s$, and 1.3 s, respectively, to fall to the ground. 

Based on the amount of time it took for a droplet to fall out of the frame of the video (10 cm) tracked using ImageJ, the equation is used to estimate the radius of droplets captured by this metrology. It is determined that this setup is able to capture droplets as small as 9.2 $\mu m$, which is below the threshold at which particles can be considered aerosols. Thus, this setup can be used to thoroughly evaluate a mask’s performance at blocking not only droplets but also aerosols.
As depicted in Figure 12, the trajectories of two different droplets are shown. The bigger droplet took 110 ms to fall 10 cm as compared to 167 ms for the small droplet. They have a calculated radius of 88 $\mu m$ and 71 $\mu m$, respectively. The trajectory of the big droplet is steeper and thus travels a shorter distance compared to the relatively smaller droplet.
\begin{figure*}[thpb]
      \centering
      \label{figure12}
      \includegraphics[scale=0.34]{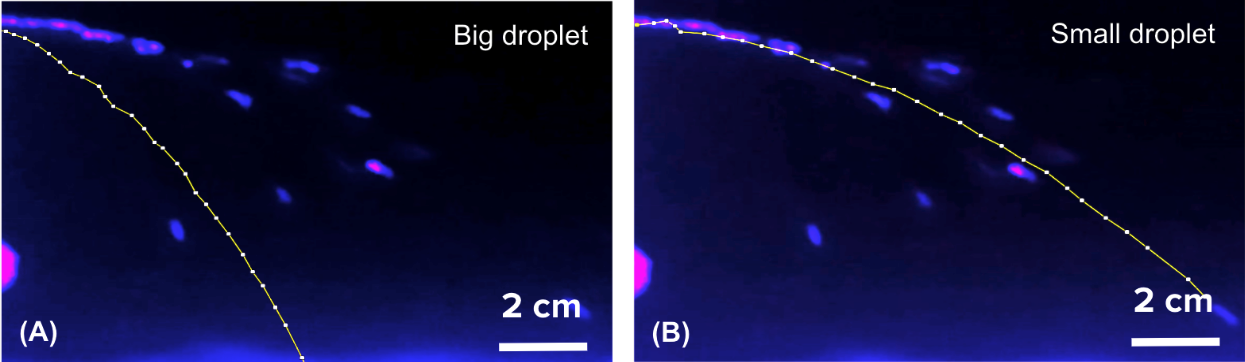}
      \caption{Trajectory of (A) a large droplet with an estimated radius of 88 $\mu m$ and (B) a small droplet with an estimated radius of 71 $\mu m$ ejected from a 8 gauge syringe needle. The paths of the droplets were tracked frame-by-frame and overlaid with the final snapshot using ImageJ.}
\end{figure*}

\subsection{Dependency of Droplet Generation and Propagation on Phonics}

Testing results on spoken words showed that the “F” and “Ph” phonics generated substantially more droplets compared to vowels and other consonants, as can be seen in Figure 13.

\begin{figure*}[thpb]
      \centering
      \label{figure13}
      \includegraphics[scale=0.485]{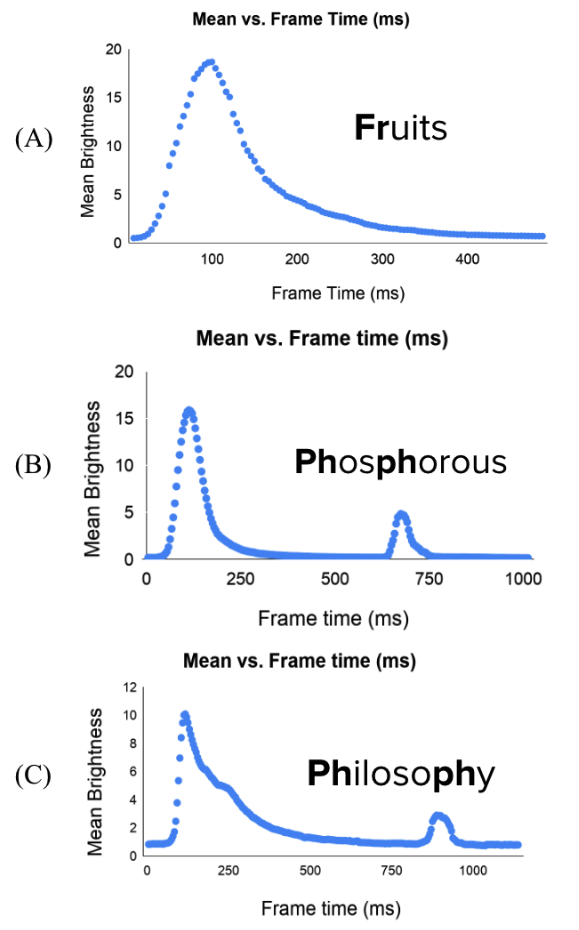}
      \caption{The mean brightness of the tonic water discharge from speaking the words (A) “Fruits”, (B) “Phosphorus”, and (C) “Philosophy” were integrated over each frame and plotted over time.}
\end{figure*}

A peak in the mean brightness of the graph indicates a sudden influx of droplets. The timestamps of these peaks correspond to whenever the “F” sound is spoken; thus, the generation of excess oral fluid droplets can be attributed to the utterance of the “F” sound. This is likely due to the vibration of the front teeth against the bottom lip while uttering these sounds propel the droplets outward. The same logic can be extended to the “Th” sound. As a result, a cloud of fluid is expelled, making these phonics more dangerous to utter in public because they can easily spread a respiratory pathogen. 

\subsection{Dependency of Droplet Generation and Propagation on Loudness}

It was also observed that the amount of fluid expelled during speech increased with the loudness levels of the sound. Figure 14A shows the superimposed mean brightness vs. time graphs for different loudness levels of speech, and the relationship between microdroplet generation and the loudness levels. Louder speech generates more droplets, as indicated by brightness levels from the data analysis. In addition, the droplets generated by louder speech linger in the air for longer than softer speech. This means that not only does louder speech result in greater microdroplet generation, but it also propels droplets further. The correlation between loudness level and the peak intensity of the recording is shown in Figure 14B. The greater the loudness level is, the more droplets are generated. 

\begin{figure*}[thpb]
      \centering
      \label{figure14}
      \includegraphics[scale=0.52]{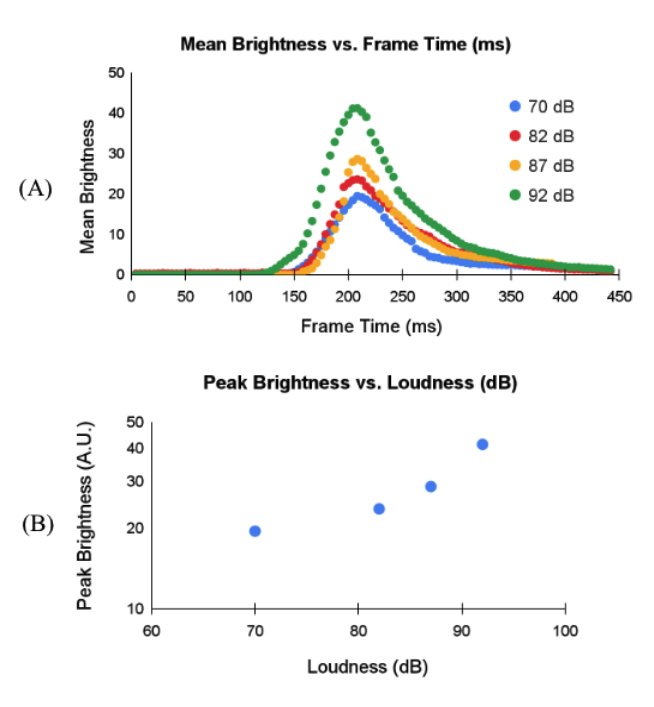}
      \caption{Graphs of (A) mean brightness over time and (B) the peak brightness values for the word “fruits” spoken at different loudness levels.}
\end{figure*}

\subsection{Dependency of Droplet Generation and Propagation on Expiratory Event}

In this section, a range of expiratory events, including speech, coughing, and sneezing were studied. As can be observed in Figure 15, the droplet cloud expelled by a sneeze lingered for much longer compared to the droplets from other expiratory events. This may be due to the variations in the aerosol to droplet ratio generated by different expiratory events. Since aerosols remain in the air for prolonged periods of time, a higher aerosol to droplet ratio would result in a much longer dissipation time. Orally generated aerosols are considered to be more dangerous than larger droplets in terms of disease spread due to their extended airborne period [35]. Aerosols can travel on air currents for up to hours before settling, during which they remain infectious. So, it can be inferred that sneezing in public is more dangerous than speaking or coughing due to the increased generation of small aerosols.

\begin{figure*}[thpb]
      \centering\vspace{1 cm}
      \label{figure15}
      \includegraphics[scale=0.48]{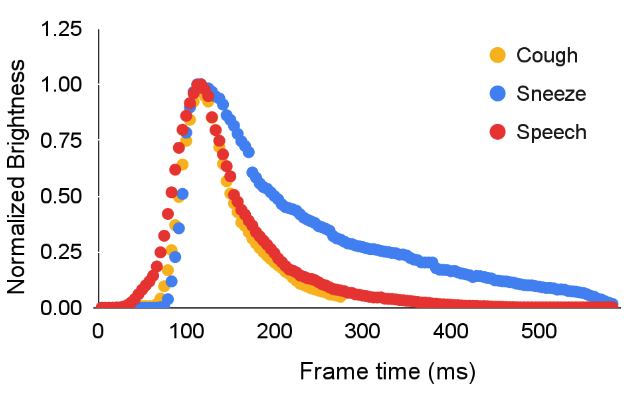}
      \caption{Three different expiratory events, including cough, sneeze, and speech, were recorded and analyzed. The normalized brightness of each event was plotted over time.}
\end{figure*}

\section{\textbf{CONCLUSIONS AND FUTURE PROSPECTS}}

In summary, a novel, home-built, low-cost, and accurate metrology to visualize oral fluid droplets and quantify mask efficacy has been developed. This fluorescence-based technique development process consists of four major blocks: setup optimization, data collection, data analysis, and applications.
In this study, droplet size and height was mathematically correlated with the amount of time it took to fall out of the frame. It is found that this prototype is capable of capturing aerosols as small as 9.2 µm.  
The proposed system can be easily built and used in a home environment and can be carried out in many different settings. The setup is easily scaleable and can be implemented into a portable prototype costing less than \$60. Compared to the original closet setup with only one UV tube light at the top, the integrated portable prototype with UV tube lights from both the top and bottom is superior in terms of illumination uniformity and intensity, thus generating data that has high intensity and low variability. Video data collected from the portable prototype displayed a 1.5-2x increase in intensity and 29-53\% less variability compared to the closet setup. Since it is easily transportable, this technique can be used for educational and mask evaluation purposes around the globe. The setup can be democratized for widespread use in the fight against contagious respiratory infections. In the future, this could also be used for studying fluid droplet and aerosol dynamics. The use of this method was simplified by automating the data processing and analysis processes to make it into a mobile application.

\vspace{1 cm}
%%%%%%%%%%%%%%%%%%%%%%%%%%%%%%%%%%%%%%%%%%%%%%%%%%%%%%%%%%%%%%%%%%%%%%%%%%%%%%%%


\begin{thebibliography}{35}
\bibitem{c1} Feehan, J. and Apostolopoulos, V. “Is COVID-19 the Worst Pandemic?” Maturitas, 149, 56-58 (2021).
\label{c1}
\bibitem{c2} Lu, J.; Gu, J.; Li, K.; Xu, C.; Su, W.; Lai, Z.; Zhou, D.; Yu, C.; Xu, B.; Yang, Z.; “COVID-19 outbreak associated with air conditioning in restaurant, Guangzhou, China,” Emerging Infectious Diseases, 26, 1628-1631 (2020).
\label{c2}
\bibitem{c3} Park, S. Y.; Kim, Y.; Yi, S.; Lee, S.; Na, B.; Kim, C. B.; Kim, J.; Kim, H. S.; Kim, Y. B.; Park, Y.; Huh, I. S.; Kim, H. K.; Yoon H. J.; Jang, H.; Kim, K.; Chang, Y.; Kim, I.; Lee, H.; Gwack, J.; Kim, S. S.; Kim, M.; Kweon, S.; Choe, Y. J.; Park, O.; Park, Y. J.; Jeong, E. K., “Coronavirus disease outbreak in call center, South Korea”, Emerging Infectious Diseases, 26, 1666-1670 (2020).
\label{c3}
\bibitem{c4} Gandhi, M.; Yokoe, D. S.; Havlir, D. V., “Editorial: Asymptomatic transmission, the Achilles’ heel of current strategies to control Covid-19,” New England Journal of Medicine, 382, 2158-2160 (2020).
\label{c4}
\bibitem{c5} Johns Hopkins University School of Medicine. Coronavirus Resource Center. https://coronavirus.jhu.edu/
\label{c5}
\bibitem{c6} World Health Organization. “Global Influenza Strategy 2019-2030,” World Health Organization. 
\label{c6}
\bibitem{c7} National Institutes of Health. “Respiratory Syncytial Virus (RSV),” NIH (2022). https://www.niaid.nih.gov/diseases-conditions/respiratory-syncytial-virus-rsv
\label{c7}
\bibitem{c8} Tang, J. W.; Li, Y.; Eames, I.; Chan, P. K. S.; Ridgway, G. L., “Factors involved in the aerosol transmission of infection and control of ventilation in healthcare premises,” Journal of Hospital Infection, 64(2), 100-114 (2006).
\label{c8}
\bibitem{c9} Tcharkhtchi, A.; Abbasnezhad, N.; Zarbini, M.; Zirak, N.; Farzaneh, S.; Chirinbayan, M., “An overview of filtration efficiency through the masks: Mechanisms of the aerosols penetration,” Bioactive Materials, 11, 6(1), 106-122 (2020).
\label{c9}
\bibitem{c10} Infection Control Today. “4 Million Deaths Caused Annually by Acute Respiratory Infections.”  (2010) https://www.infectioncontroltoday.com/view/4-million-deaths-caused-annually-acute-respiratory-infections
\label{c10}
\bibitem{c11} Cook, T. M, “Personal protective equipment during the coronavirus disease (COVID) 2019 pandemic- a narrative review,” Anaesthesia, 75(7), 920-927 (2020).
\label{c11}
\bibitem{c12} Burki, T., “Global shortage of personal protective equipment,” The Lancet: Infectious Diseases, 20(7), 785-786 (2020).
\label{c12}
\bibitem{c13} Alexandar, G., “Can Face Masks be Recycled?” Earth911, (2021). https://earth911.com/living-well-being/can-face-masks-be-recycled/
\label{c13}
\bibitem{c14} Ma, J.; Chen, F.; Zhang, Z.; Li, Y.; Liu, J.; Chen, C.; Pan, K., “Eukaryotic community succession on discarded face masks in the marine environment,” Science of The Total Environment, 854, 15882, (2023).
\label{c14}
\bibitem{c15} Sangkham, S., “Face mask and medical waste disposal during the novel COVID-19 pandemic in Asia,” Case Studies in Chemical and Environmental Engineering, 2, 100052 (2020).
\label{c15}
\bibitem{c16} Bhaik, A.; Singh, V.; Gandotra, E.; Gupta, D., “Detection of improperly worn face masks using deep learning – a preventive measure against the spread of COVID-19,” International Journal of Interactive Multimedia and Artificial Intelligence, 7(7) 14-25 (2021).
\label{c16}
\bibitem{c17} Centers for Disease Control and Prevention. “Guidance for Wearing Masks,” (2021). https://www.cdc.gov/coronavirus/2019-ncov/prevent-getting-sick/cloth-face-cover-guidance.html
\label{c17}
\bibitem{c18} Bourouiba, L., “A sneeze,” New England Journal of Medicine, 375 (8), e15 (2016).
\label{c18}
\bibitem{c19} Christopherson, D. A.; Yao, W. C.; Lu, M.; Vijayakumar, R.; Sedaghat; A. R., “High-efficiency particulate air filters in the era of COVID-19: Function and efficacy.” SAGE journals, 163(6),  1153-1155 (2020).
\label{c19}
\bibitem{c20} Lakowicz, J. R., “Introduction to fluorescence. Principles of Fluorescence Spectroscopy.” Pg. 1-23. (1999).
\label{c20}
\bibitem{c21} Anfinrud, P.; Bax, C.; Bax, A.; “Visualizing Speech-Generated Oral Fluid Droplets with Laser Light Scattering.” New England Journal of Medicine, 382, 2061-2063 (2020).
\label{c21}
\bibitem{c22} Fry, T. R., “Laser Safety.” Veterinary Clinics: Small Animal Practice, 32(3), 535-547 (2002).
\label{c22}
\bibitem{c23} Baskar, P., “Coronavirus FAQ: What Does It Mean If I Can Blow Out A Candle While Wearing A Mask?” NPR, (2020).
\label{c23}
\bibitem{c24} Mythbusters. https://www.youtube.com/watch?v=3vw0hIs2LEg
\label{c24}
\bibitem{c25} Yohalem, S. B., “Quinine in tonic water.” Journal of the American Medical Association. 153(14), 1304 (1953).
\label{c25}
\bibitem{c26} Lawson-Wood, K. and Evans, K., “Determination of Quinine in Tonic Water Using Fluorescence Spectroscopy.” PerkinElmer, (2018). https://www.perkinelmer.com/lab-solutions/resources/docs/APP\_Quinine\_in\_Tonic\_Water\_014133\_01.pdf
\label{c26}
\bibitem{c27} Ramsay, W. and Dobbie, J. J., “On the decomposition-products of quinine.” Journal of the Chemical Society, Transactions, 33, 102-104 (1878).
\label{c27}
\bibitem{c28} University of Rochester, “Ultraviolet Light Safety Guidelines.” University of Rochester (2021). <https://www.safety.rochester.edu/ih/uvlight.html>
\label{c28}
\bibitem{c29} VideoLAN. VLC Media Player. https://www.videolan.org/
\label{c29}
\bibitem{c30} National Institutes of Health. ImageJ. https://imagej.nih.gov/ij/
\label{c30}
\bibitem{c31} Malloy, J.; Quintana, A.; Jensen, CJ.; Liu, K., “Copper foam as a highly efficient, durable filter for reusable masks and air cleaners,” Nano Letters, 21, 7, 2968–2974 (2021).
\label{c31}
\bibitem{c32} Jayaweera, M.; Perera, H.; Gunawardana, B.; Manatunge, J., “Transmission of COVID-19 virus by droplets and aerosols: A critical review on the unresolved dichotomy.” NCBI, 188, 109819 (2020).
\label{c32}
\bibitem{c33} Wang, H. P.; Li, Z. B.; Zhang, X. L.; Zhu, L. X.; Liu, Y.; Wang, S. Z., “The motion of respiratory droplets produced by coughing.” Physics of Fluids, 32, 1251-1252 (2020). 
\label{c33}
\bibitem{c34} Netz, R. R.; Eaton, W. A.; “Physics of virus transmission by speaking droplets.” PNAS, 117(41), 25209-25211 (2020).
\label{c34}
\bibitem{c35} Penn Medicine, “COVID-19: Droplet or Airborne Transmission?” Penn Medicine Epidemiologists Issue Statement. Penn Medicine Physician Blog (2020). https://www.pennmedicine.org/updates/blogs/penn-physician-blog/2020/august/airborne-droplet-debate-article\# 
\label{c35}


\end{thebibliography}
\end{document}